\documentclass{article} % For LaTeX2e
\usepackage{iclr2027_conference,times}

\usepackage{amsmath,amsfonts,bm}

\def\eqref#1{equation~\ref{#1}}
\def\1{\bm{1}}

\DeclareMathAlphabet{\mathsfit}{\encodingdefault}{\sfdefault}{m}{sl}
\SetMathAlphabet{\mathsfit}{bold}{\encodingdefault}{\sfdefault}{bx}{n}

\usepackage{hyperref}
\usepackage{times}
\usepackage{soul}
\usepackage{url}
\usepackage[utf8]{inputenc}
\usepackage{multirow}
\usepackage[small]{caption}
\usepackage{graphicx}
\usepackage{amsmath}
\usepackage{amsthm}
\usepackage{booktabs}
\usepackage{algorithm}
\usepackage{algorithmic}
\usepackage[switch]{lineno}
\usepackage{latexsym}
\usepackage{makecell}
\usepackage{float}
\usepackage{svg}
\usepackage{afterpage}
\usepackage{amsmath}
\usepackage{amssymb}
\usepackage{float}

\title{Latent Prediction Yields Cross-Subject Alignable Representations in Sensory-Evoked fMRI}

\author{Serwan Jassim, Victoria Bosch, Daniel Anthes, Elia Bruni\textsuperscript{*} \& Tim Kietzmann\textsuperscript{*} \\
Institute of Cognitive Science\\
University of Osnabrück\\
\texttt{\{serwan.jassim, victoria.bosch, daniel.anthes,}\\
\texttt{elia.bruni, tim.kietzmann\}@uos.de}
}

\iclrfinalcopy % Uncomment for camera-ready version, but NOT for submission.
\begin{document}
\renewcommand{\headrulewidth}{0pt}

\maketitle
\begingroup
\renewcommand{\thefootnote}{*}
\footnotetext{Elia Bruni and Tim Kietzmann share senior authorship.}
\endgroup

\begin{abstract}
Decoding sensory-evoked functional magnetic resonance imaging (fMRI) data is a central analytical tool in cognitive neuroscience. Here we ask whether self-supervised learning (SSL), which has been scarcely explored in this domain, can yield representations that support strong downstream performance. To do so, we adapt a joint embedding predictive architecture (JEPA) to train participant-specific encoders on brain data alone. The encoders never see the stimulus images or any representation derived from them. Applied to the Natural Scenes Dataset (NSD), we evaluate our learned brain embeddings on the downstream tasks of brain captioning and image reconstruction, as well as cross-participant generalization. Our self-supervised approach achieves competitive performance with parameter-rich supervised pipelines that are trained end-to-end with access to image and language data. We further find that encoders trained independently for different participants can be aligned fully unsupervised, using neither images nor stimuli shared between them. After alignment, a neural decoding head trained on one participant can be applied to another without fine-tuning and retains most of its within-subject performance on image reconstruction and brain captioning. By removing the need for images and shared stimuli, our approach is a step towards pooling fMRI data across participants and, eventually, across datasets.
% Second, we find that our encoders can be aligned across participants in a fully unsupervised fashion, again without access to images or shared stimuli. Despite being a challenging task due to anatomical, functional, and cognitive variability of the participants, our approach achieves performance levels previously only observed for models that rely either on shared stimuli across subjects, or on anchoring representations to pre-trained vision encoders. By removing the need for images and shared stimuli, our approach thus paves the way towards scaling representation learning to many fMRI datasets.
\end{abstract}

\section{Introduction}
The decoding of information from brain activity is a central analytical tool in cognitive neuroscience. Classical applications ask whether experimental conditions, visual categories, or other stimulus properties can be read out from measured brain responses alone. More recently, advances in generative modeling and the availability of large-scale neural datasets, such as the Natural Scenes Dataset (NSD)~\citep{allen2022massive}, have substantially expanded the scope of which information can be derived from brain data. Capabilities of brain decoding now include largely veridical image reconstruction from functional MRI (fMRI) responses, obtained while participants observe visual stimuli~\citep{takagi2023high, Ozcelik2023BrainDiffuserNS, mindeye2}, up to approaches that map brain activity into language representations, enabling versatile brain captioning and question-answering interfaces beyond reconstruction itself~\citep{mai2023unibrain, bosch2025brainlanguage}. 

These approaches, however, place substantial demands on the available data. Modern decoding pipelines are typically trained separately for each participant, and many additionally rely on paired supervision between brain responses and stimulus-derived representations, such as image embeddings or language annotations. Both requirements are restrictive. Even unusually large fMRI datasets contain orders of magnitude fewer training examples per participant than are commonly used to train modern machine-learning models, simply because acquiring fMRI data is time-consuming and expensive. Pooling data across participants could substantially increase the amount of brain data available for representation learning, but this is difficult because brains differ in functional organization and in the precise spatial pattern through which information is represented. Consequently, a decoder trained on one participant does not straightforwardly generalize to another. Existing approaches to functional alignment can reduce this problem, but typically depend on paired measurements, i.e., responses from multiple participants to the same stimuli.

An appealing alternative would therefore be to learn brain representations that address both limitations at once: such a representation should make effective use of the brain data itself, without requiring large-scale external supervision during representation learning, while at the same time enabling versatile downstream applications and exposing a structure that can be aligned across participants despite their anatomical and functional differences. Self-supervised learning (SSL) provides a natural framework for this objective. Rather than learning brain representations by predicting an image embedding or a language annotation, we learn them from the structure that the measured brain responses share across repetitions of the same stimulus. Because the objective is self-supervised, the resulting representation is not tied to the feature space of a particular external image or language model, and can serve multiple readouts. In addition, the objective does not depend on the stimulus modality: it applies to any experiment that repeats its stimuli.

Here, we explore this possibility for stimulus-evoked fMRI using a joint-embedding predictive architecture (JEPA). SSL has previously been applied to fMRI, particularly in the resting-state setting for downstream phenotype or clinical prediction~\citep{ortegacaro2023brainlm, dong2024brainjepa}. We instead ask whether it can recover representations of stimulus-evoked activity that retain the fine-grained information required for generative decoding. We pre-train one encoder per participant on NSD, asking it to predict the latent representation of masked cortical parcels from a different repetition of the same stimulus. Pairing repetitions this way leaves stimulus-locked structure as the only predictable signal and does not reward the model for measurement-specific noise; the model also never sees the stimulus, an image embedding, or a language annotation during pre-training.

We find that the resulting brain embeddings are highly expressive. After self-supervised pre-training, we freeze the encoders and train downstream heads for image reconstruction and brain captioning using MindEye2~\citep{mindeye2} and CorText~\citep{bosch2025brainlanguage}, respectively. Despite the absence of image or language supervision during encoder training, these brain embeddings support downstream performance competitive with the corresponding pipelines trained end-to-end with stimulus supervision. Thus, much of the information required for mapping brain activity into modern image and language models can be captured through latent prediction directly within brain data, reducing the amount of task-specific parameter learning required at the decoding stage.

Self-supervised training also yields representations that can be aligned across participants without externally defined anchors, such as shared stimuli or stimulus content. This has recently been shown for compact, largely linear embeddings of NSD participants~\citep{MarcosManchn2026PlatonicRI}. What remains open is whether alignability survives in representations rich enough to drive generative decoders and whether a decoder trained on one participant works on the aligned representations of another. Applying that alignment procedure, extended to parcel sequences, to our per-participant encoders, we find the full parcel embeddings still related by orthogonal maps recoverable without paired data, and a decoding head trained on one participant applies unchanged to another's aligned representations, retaining most of its within-participant quality. Self-supervised pre-training therefore provides both an expressive within-participant representation and a basis for cross-participant transfer.

% More surprisingly, self-supervised training also produces representations that offer themselves to be aligned across participants, without the need for externally defined alignment anchors, such as shared stimuli or latent stimulus embeddings. In particular, independently trained participant-specific encoders exhibit representational geometries that can be brought into correspondence using an unsupervised alignment procedure extending previous work on geometric alignment of embedding spaces~\citep{conneau2018word, jha2025harnessing, MarcosManchn2026PlatonicRI}. Crucially, estimating this correspondence requires neither access to the original images nor stimuli shared between participants. Once aligned, a decoding head trained on one participant can be applied directly to the representations of another participant while retaining a substantial fraction of within-participant performance. This feature is not observed in baseline brain data prior the self-supervised embedding. Self-supervised pre-training therefore provides both an expressive within-participant representation and a basis for cross-participant transfer that can be harnessed for downstream applications and for deriving foundation models for brain science.

Our contributions are:
\begin{itemize}
\item We introduce self-supervised pre-training of stimulus-evoked fMRI encoders using a brain-only JEPA objective. The objective exploits repeated responses to the same stimulus, but requires no access to the stimulus itself or to externally pre-trained image or language representations.
\item We show that the resulting frozen brain embeddings simultaneously support both image reconstruction and brain captioning at performance levels competitive with corresponding end-to-end supervised pipelines, despite requiring substantially less task-specific parameter learning.
\item We show that independently trained, participant-specific brain embeddings can be aligned fully unsupervised, without images or stimuli shared across participants, enabling cross-participant decoding without retraining the downstream decoder.
\end{itemize}

Together, these results suggest that latent prediction can provide a common representational basis for both expressive brain decoding and cross-participant generalization. By decoupling representation learning from stimulus-derived supervision and cross-participant alignment from shared stimuli, this approach opens a path towards combining brain data across participants and, ultimately, across datasets that were not originally acquired with joint decoding in mind.
\section{Related Work}
\paragraph{Self-Supervised Representation Learning.} SSL is a family of training objectives that require only input data and no target labels. It is commonly used to pre-train representations that can be deployed across a range of downstream tasks. Prominent SSL objectives include reconstruction or masked prediction in the input space~\citep{vincent2008extracting, pathak2016context, devlinetal2019bert, He_2022_CVPR, xie2022simmim}, and prediction or matching in latent representation spaces~\citep{chen2020simple, grill2020byol, caron2021emerging, bardes2022vicreg}. Joint embedding predictive architectures (JEPAs) \citep{lecun2022path, assran2023selfsupervised, bardes2024vjepa, assran2025vjepa2} are a more recent class of latent-prediction models: rather than reconstructing raw observations, they predict latent representations of masked regions from a visible context. Because unpredictable input-level details need not be represented faithfully~\citep{littwin2024jepa}, this objective may be particularly suitable for fMRI, where scanner, physiological, and other non-stimulus-related variation is common.  

SSL methods have been widely applied in neuroimaging~\citep{kim2023swift, ortegacaro2023brainlm, yang2024brainmass, cortexmae2025, wang2026neurostorm, omnifmri2026}, though many operate on resting-state data and are evaluated on phenotype and clinical readouts. In that setting, \citet{dong2024brainjepa} report that a JEPA objective outperforms reconstruction- and contrastive-based SSL on demographic and disease prediction, motivating our use of this objective family for sensory-evoked fMRI. Closest to our setting, \citet{MarcosManchn2026PlatonicRI} construct subject-specific representations on NSD~\citep{allen2022massive} using a largely linear pipeline (PCA and multiset CCA), refine them with a contrastive objective, and evaluate them through cross-trial retrieval. We differ by learning these representations end-to-end and evaluating them on generative readouts rather than retrieval alone.

\paragraph{Decoding from fMRI: vision and language.}
Advances in generative architectures for vision and language have enabled increasingly detailed decoding of sensory-evoked fMRI. One line of work maps brain signals into the latent space of generative image models to reconstruct the viewed stimulus~\citep{takagi2023high, Ozcelik2023BrainDiffuserNS, Chen_2023_CVPR, scotti2023minds_eye}; a parallel line maps them into language models to generate textual descriptions instead~\citep{ferrante2023brain, mai2023unibrain, chen2023mindgpt, huang2024brainchat, pmlr-v267-qiu25e}. We adopt MindEye2~\citep{mindeye2} for image reconstruction and CorText~\citep{bosch2025brainlanguage} for captioning. Both methods train the brain-side encoder against stimulus-derived targets, such as frozen image embeddings or caption tokens. We instead use these pipelines as downstream readouts --- trained against the same targets --- of a frozen brain encoder whose pre-training objective has no access to stimulus images or stimulus-derived representations.

\paragraph{Cross-Subject Functional Alignment.}
Brains differ across individuals in both anatomy and functional organization, so the same measurement location does not correspond across people. Because establishing that correspondence is difficult, decoding models are typically built separately for each individual. Hyperalignment addressed this by learning transformations from responses to shared stimuli~\citep{haxby2011common}, an approach extended by the shared response model~\citep{sharedresponsemodeling} and optimal-transport formulations~\citep{optimaltransport} and inherited by several deep learning-based methods~\citep{chen2016convolutionalautoencodermultisubjectfmri, huang2021learningsharedneuralmanifolds}. More recent methods remove the requirement for shared stimuli, but instead anchor participants to representations of pre-trained image encoders \citep{WASSERMAN2026121741, papathanasiou2026task}. Our setting removes both requirements: neither pre-training nor alignment uses shared stimuli or representations derived from the stimulus images.

Outside neuroimaging, it has already been shown that different embedding spaces can be brought into correspondence without paired examples by matching their internal similarity structure~\citep{conneau2018word, alvarezmelisjaakkola2018gromov, jha2025harnessing}. \citet{MarcosManchn2026PlatonicRI} recently adopted this framework for sensory-evoked fMRI data, aligning per-participant embeddings without paired stimuli across participants or access to stimulus images. We use their alignment procedure to test whether independently and end-to-end learned neural representations develop corresponding geometries.
\section{Method}\label{sec:method}
Our method has three components: we pre-train one JEPA encoder per participant on parcellated sensory-evoked fMRI responses (Section~\ref{subsec:pretraining}), evaluate the resulting brain embeddings with generative decoding heads trained on top of the frozen encoder (Section~\ref{subsec:decoding}), and align the independently trained encoders across participants without shared stimuli (Section~\ref{subsec:alignment}). Appendix~\ref{app:training_params} lists the parameters for all three. We first describe the data and the parcel representation that all three build on.

\subsection{Dataset}\label{subsec:dataset}
We use NSD~\citep{allen2022massive} for both pre-training and downstream evaluation. NSD is a sensory-evoked 7T fMRI dataset in which eight participants viewed up to 10{,}000 images from MS COCO~\citep{lin2014microsoft} over 40 scanning sessions, where each image was presented up to three times in total. We restrict our experiments to participants 1, 2, 5, and 7, the four who completed all scanning sessions. For each participant, the training set comprises the 9{,}000 unique stimuli seen only by that participant, while the 1{,}000 shared stimuli, which all participants viewed, are held out for evaluation. We use the fsaverage-surface betas provided with NSD and z-score each session independently within each participant, with train and test sets standardized separately. We then parcellate the surface data with the 500-parcel 7-network Schaefer atlas~\citep{schaefer2018local}, yielding 500 parcels per trial.

We explore restricting the readout to a subset of parcels containing visually responsive regions. We define this subset anatomically and independently of any functional response. \citet{Glasser2016} group their 180 cortical areas into 22 coarser cortical divisions; we assign each Schaefer parcel to the division containing the largest share of its vertices on the shared fsaverage surface, and retain the parcel tokens assigned to the five visual divisions --- Primary Visual, Early Visual, Dorsal Stream Visual, Ventral Stream Visual, and MT+ Complex. This yields 66 parcels spanning 41{,}248 of the 299{,}761 labeled vertices (13.8\%), 35 in the left and 31 in the right hemisphere. Because the two atlases' boundaries do not coincide, individual Schaefer parcels may span multiple Glasser divisions; nonetheless, 44 of the 66 selected parcels lie entirely within visual divisions, and, on average, 93.1\% of a selected parcel's vertices do. Since the criterion uses only anatomical atlas overlap, the subset is identical across participants and involves no selection on the neural data.

\begin{figure*}[ht]
\centering
\includegraphics[width=\textwidth]{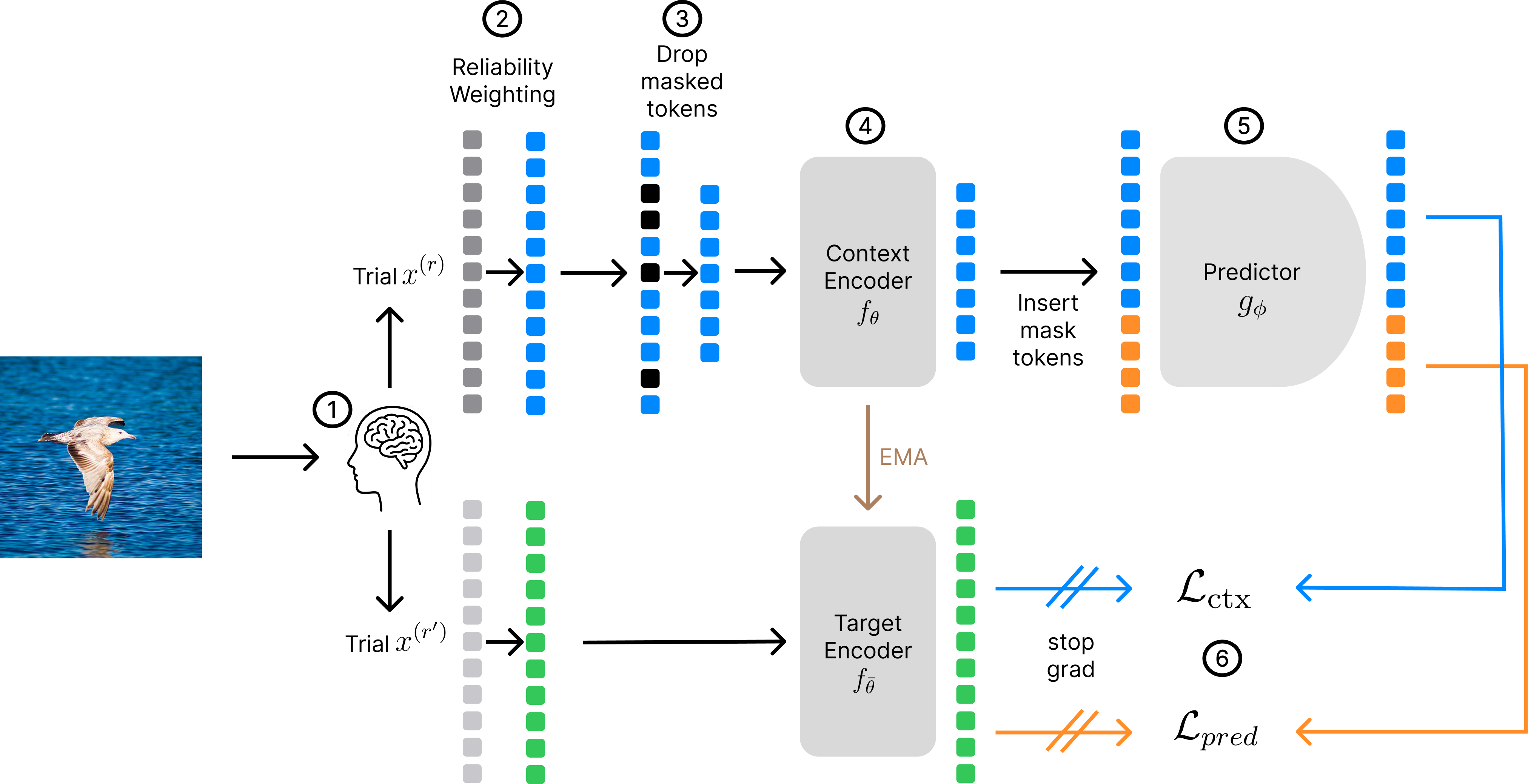}
\caption{Overview of our self-supervised brain encoder training. \textbf{(1)} A subject's brain response to a stimulus image is recorded across repeated presentations; we sample two of the three repetitions, $r$ and $r'$, one for the context encoder path and one for the target encoder path. After Schaefer parcellation, each trial's brain response is represented as a sequence of parcels. \textbf{(2)} Reliability weights are applied to the raw vertices within each parcel. \textbf{(3)} Mask positions are sampled, and the corresponding tokens are dropped from the context encoder input sequence. \textbf{(4)} The context encoder $f_\theta$ produces context representations using per-parcel MLPs, followed by transformer layers. The target encoder $f_{\bar{\theta}}$, updated as an exponential moving average of the context encoder, encodes the unmasked complementary trial; no gradient flows through this path. \textbf{(5)} Using learnable mask tokens, the predictor $g_{\phi}$ predicts the missing positions from the context representations. \textbf{(6)} Predictor outputs at masked positions, as well as at context positions, are supervised against the corresponding target representations under an $L_2$ loss.}
\label{fig:pipeline}
\end{figure*}

\subsection{Self-Supervised Brain Encoder}\label{subsec:pretraining}
The functional parcellation described above maps naturally onto the JEPA setup, which learns by predicting the representations of masked tokens --- typically image or video patches, and here the parcels of a single trial's brain response. Figure~\ref{fig:pipeline} gives an overview of the training process.

Let $x \in \mathbb{R}^{V}$ denote the response of a single trial, where $V$ is the number of fsaverage vertices covered by the atlas. The parcellation partitions these vertices into $P = 500$ disjoint parcels, so that $x = (x_1, \dots, x_P)$ with $x_p \in \mathbb{R}^{V_p}$ and $\sum_{p=1}^{P} V_p = V$. Within each parcel, vertices are weighted by their reliability --- how consistently a vertex responds across repeated presentations of the same stimulus --- so that noise-dominated vertices contribute less: $\tilde{x}_p = w_p \odot x_p$, with $w_p \in \mathbb{R}^{V_p}$ computed following \citet{MarcosManchn2026PlatonicRI} and normalized to sum to one.

The context encoder $f_\theta$ begins with a parcel embedding layer. Because the number of vertices $V_p$ differs across parcels, it comprises a two-layer MLP $e_p: \mathbb{R}^{V_p} \rightarrow \mathbb{R}^{D}$ per parcel. Together with learnable positional embeddings $E \in \mathbb{R}^{P \times D}$, it maps a brain response of $V$ vertices to a sequence of $P$ tokens of width $D$. We use learnable rather than structured positional embeddings, as we do not want to impose a neighborhood structure on the parcel sequence. A stack of transformer blocks then operates on this sequence. The predictor $g_\phi$ is a transformer at width $D_p$ and predicts masked tokens using learnable mask tokens. Following the standard JEPA procedure, we initialize a target encoder $f_{\bar{\theta}}$ from the context encoder and update it as an exponential moving average of its weights.

Unlike image patches, contiguous blocks of the parcel sequence do not correspond to anatomically neighboring cortical regions, so we sample random token sets rather than the contiguous blocks commonly used in JEPA pre-training. Masking randomly samples one context set $C \subset \{1, \dots, P\}$ and $M$ target sets $T_1, \dots, T_M \subset \{1, \dots, P\}$. We sample context and target sets independently, and remove tokens appearing in any target set from the context, such that $C \cap T_m = \emptyset$ for all $m$.

To encourage the learned brain embeddings to discard stimulus-independent information, the context and target encoders receive \emph{different repetitions of the same stimulus}. Two repetitions share the stimulus but differ in scanner noise, physiological state, and session effects. Since these components are independent across presentations, they carry no information about the target representation, and the objective therefore gives the encoder no incentive to represent them; only stimulus-related information is predictable from one repetition to the other. Let $x^{(r)}$ and $x^{(r')}$ be two repetitions of one image, sampled without
replacement from the three available and resampled each epoch. Given the context representations of $\tilde{x}^{(r)}$, the predictor produces the sequence
\begin{equation*}
    \hat{z} = g_\phi\big(f_\theta(\tilde{x}^{(r)}, C),\, T_m\big) =  \big[\hat{z}_C \mid \hat{z}_{T_m}\big]
\end{equation*}
while the target encoder produces
\begin{equation*}
    z = f_{\bar{\theta}}\big(\tilde{x}^{(r^\prime)}\big)
\end{equation*}
for the entire unmasked trial $\tilde{x}^{(r^\prime)}$. Following V-JEPA 2.1~\citep{murlabadia2026vjepa2_1}, we add a loss term over context predictions, making our per-sample loss
\begin{equation*}
    \mathcal{L} = \mathcal{L}_{\text{ctx}} + \mathcal{L}_{\text{pred}}
    = \frac{1}{\lvert C \rvert}\big\lVert \hat{z}_{C} - z_C \big\rVert_2^2
    + \frac{1}{M} \sum_{m=1}^M \frac{1}{\lvert T_m \rvert} \big\lVert \hat{z}_{T_m} - z_{T_m} \big\rVert_2^2.
\end{equation*}

\subsection{Generative Neural Decoding}\label{subsec:decoding}
To evaluate the quality of our pre-trained encoder, we train generative decoding heads on top of the resulting brain embeddings. Throughout, we keep the encoder frozen: its weights are never updated by any task-specific learning signal. All readouts use the target encoder from the pre-training stage.

\paragraph{Image Reconstruction.} The first task is image reconstruction, where the goal is to generate the image a participant viewed while their brain response was recorded. We adopt the MindEye2~\citep{mindeye2} pipeline, which uses pre-trained CLIP and VAE image representations as supervision targets for its decoder. Its head consumes a $256 \times 1664$ token sequence produced by the MindEye2 backbone. To bridge into this space from our brain embeddings of shape $P \times D$, we apply an attentive pooling layer with $256$ queries followed by a linear map to $1664$ dimensions. The pipeline trains a contrastive projection alongside the reconstruction objective, from which we additionally report retrieval: \emph{image} retrieval ranks the correct image among a set of candidates given a brain response, whereas \emph{brain} retrieval ranks the correct brain response given an image. In contrast to MindEye2, we do not pre-train the head on other subjects. The inputs also differ: MindEye2 operates on the participant-specific \texttt{nsdgeneral} region, defined by the NSD authors as the voxels most responsive to visual stimuli, whereas our encoders read the Schaefer parcellation.

\paragraph{Brain Captioning.} The second task is decoding textual descriptions of the stimulus images from the brain embeddings. We use the CorText~\citep{bosch2025brainlanguage} framework, which integrates brain data into a pretrained LLM, enabling open-ended, natural-language decoding akin to visual question-answering setups. Most LLM-based methods pair a brain encoder with a vision-language model and train the encoder against the representations of that model's vision encoder, so that image representations act as supervision targets. CorText instead trains its brain encoder end-to-end from brain responses alone, under a cross-entropy loss over target answer tokens sourced from VQA datasets defined over COCO images (see \citet{bosch2025brainlanguage} for more information). Our pre-trained encoders are therefore a drop-in replacement for its brain encoder, with no image supervision entering the pipeline at any point. We project our brain embeddings into the representation space of the pre-trained LLM (Llama 3.1 8B-Instruct) with a projection layer consisting of a two-layer MLP, following common practice in vision-language models~\citep{Liu_2024_CVPR}. Training proceeds in two stages: first, only the projection layer is trained; second, the LLM is additionally tuned with low-rank adaptation~\citep{hu2022lora}. The pre-trained brain encoder remains frozen in both stages.

\subsection{Cross-Subject Alignment}\label{subsec:alignment}
To align our independently pre-trained brain encoders across participants, we employ the alignment scheme of \citet{MarcosManchn2026PlatonicRI}. The method matches the internal similarity structure of two embedding spaces without paired samples: it summarizes each space with $k$-means centroids, matches centroids across spaces by solving a quadratic assignment problem on their similarity structure, and uses the resulting correspondence to fit an orthogonal map via Procrustes transformation. This initial map is then refined by iterative closest point, which alternates between re-estimating pseudo-pairs under the current map and refitting it. All steps use only the representations of each participant's brain responses from the training set; no stimulus correspondence or label information enters at any point.

While the pipeline of \citet{MarcosManchn2026PlatonicRI} produces a single $128$-d embedding per participant, our encoders produce a sequence of $P$ tokens of width $D$, with $PD \gg 128$. Learning a single map $R \in \mathbb{R}^{PD \times PD}$ is infeasible, so we instead fit one orthogonal map per token, $R^{(t)} \in \mathbb{R}^{D \times D}$. We whiten the representations with each participant's own training statistics — a single ZCA transform shared across tokens rather than a per-token one — which improves alignment quality (see Table~\ref{tab:cross_subject_retrieval}). Writing $W_\text{A}$ and $\mu_\text{A}$ for participant $A$'s whitening matrix and mean, a token $t$ of $A$'s brain embedding is mapped into $B$'s frame as
\begin{equation}\label{eq:alignment}
    z^{(t)}_{\text{A} \rightarrow \text{B}}
    = W_\text{B}^{-1}\Big( W_\text{A}\big(z^{(t)}_{\text{A}} - \mu_\text{A}\big)
      R^{(t)}_{\text{A} \rightarrow \text{B}} \Big) + \mu_\text{B}.
\end{equation}
Both $W$ and $\mu$ are estimated from each participant's own training representations, so the alignment remains free of cross-participant supervision.

We evaluate alignment quality using cross-participant retrieval: given one participant's representation of a stimulus, the task is to retrieve another's representation of the same stimulus. We further evaluate cross-participant decoding: $A$'s representations are mapped into $B$'s frame, where $B$'s decoding head reads them without retraining.
\section{Results}

\begin{table}[t]
\centering
\small
\setlength{\tabcolsep}{3pt}
\caption{
Caption generation from brain activity on the NSD test set, averaged across subjects 1, 2, 5, and 7. Higher is better for all metrics; the best result in each column and group (within-subject vs.\ cross-subject) is shown in bold. All variants of our method use frozen pre-trained brain encoders within the CorText pipeline. \emph{Visual readout} encodes all 500 parcels but exposes only visual cortex tokens to the decoding head, while \emph{Full sequence} uses all 500 encoded parcels. For cross-subject decoding, a head trained on one subject is applied to the aligned representations of another, averaged over all 12 ordered subject pairs. Control conditions remove stimulus correspondence, brain input, or cross-subject alignment, respectively. ${\dagger}$ CorText is trained fully supervised.
}
\label{tab:cortext_caption}

\begin{tabular}{lccccccccc}
\toprule
Method
& \multicolumn{4}{c}{BLEU}
& METEOR
& ROUGE-L
& CIDEr
& CLIP-S
& RefCLIP-S \\
\cmidrule(lr){2-5}
& 1 & 2 & 3 & 4 & & & & & \\
\midrule

CorText$^{\dagger}$
& \textbf{0.628}
& \textbf{0.440}
& 0.300
& 0.204
& \textbf{0.202}
& 0.462
& \textbf{0.679}
& \textbf{0.649}
& \textbf{0.716} \\

\midrule
\multicolumn{10}{l}{\emph{Ours: within-subject}} \\
\quad Full sequence
& 0.625
& 0.438
& 0.303
& \textbf{0.212}
& 0.201
& \textbf{0.464}
& 0.673
& 0.635
& 0.705 \\

\quad Visual readout
& 0.623
& 0.439
& \textbf{0.304}
& \textbf{0.212}
& 0.200
& 0.463
& \textbf{0.674}
& 0.633
& 0.703 \\
\multicolumn{10}{l}{\emph{Control, within-subject}} \\
\quad Trial permutation
& 0.391
& 0.176
& 0.078
& 0.039
& 0.090
& 0.296
& 0.061
& 0.394
& 0.474 \\

\quad Zeroed betas
& 0.190
& 0.078
& 0.028
& 0.010
& 0.094
& 0.193
& 0.001
& 0.364
& 0.428 \\

%\quad Visual input
%& 0.608
%& 0.419
%& 0.283
%& 0.194
%& 0.194
%& 0.451
%& 0.639
%& 0.629
%& 0.697 \\

\midrule
\midrule
\multicolumn{10}{l}{\emph{Ours, cross-subject}} \\
\quad Full sequence
& 0.500
& 0.317
& 0.197
& 0.127
& 0.159
& 0.396
& 0.394
& 0.568
& 0.639 \\
\quad Visual readout
& \textbf{0.549}
& \textbf{0.359}
& \textbf{0.233}
& \textbf{0.155}
& \textbf{0.168}
& \textbf{0.416}
& \textbf{0.453}
& \textbf{0.579}
& \textbf{0.651} \\
\multicolumn{10}{l}{\emph{Control, cross-subject}} \\
\quad \makecell[l]{Visual-readout\\w/o alignment}
& 0.038
& 0.006
& 0.000
& 0.000
& 0.017
& 0.046
& 0.000
& 0.503
& 0.569 \\

\bottomrule
\end{tabular}
\end{table}

\begin{table}[t]
\centering
\small
\setlength{\tabcolsep}{2pt}
\caption{Image reconstruction from brain activity on the NSD test set, scores averaged across subjects 1, 2, 5, and 7. Arrows indicate the direction of improvement; the best result in each column and group (within-subject vs.\ cross-subject) is shown in bold. All variants of our method use frozen pre-trained brain encoders within the MindEye2 pipeline. \emph{Visual readout}, \emph{Full sequence}, the cross-subject protocol, and the control conditions are defined as in Table~\ref{tab:cortext_caption}. Retrieval uses sets of 300 candidates sampled from the full test set, with scores averaged over 30 independent draws. Brain retrieval is undefined for the zeroed-betas control, where every trial yields the same embedding, and is therefore omitted. ${\dagger}$ MindEye2 is trained fully supervised and uses all other subjects to pre-train the decoder.}
\label{tab:image_reconstruction}
\begin{tabular}{lcccccccccc}
\toprule
Method & \multicolumn{4}{c}{Low-Level} & \multicolumn{4}{c}{High-Level}
       & \multicolumn{2}{c}{Retrieval} \\
\cmidrule(lr){2-5} \cmidrule(lr){6-9} \cmidrule(lr){10-11}
  & PixCorr$\uparrow$ & SSIM$\uparrow$ & Alex(2)$\uparrow$ & Alex(5)$\uparrow$
 & Incep$\uparrow$ & CLIP$\uparrow$ & Eff$\downarrow$ & SwAV$\downarrow$
 & Image$\uparrow$ & Brain$\uparrow$ \\
\midrule
MindEye2$^{\dagger}$ & \textbf{0.278} & 0.328 & 95.2 & \textbf{99.0} & \textbf{96.4} & \textbf{94.5} & \textbf{0.622} & \textbf{0.343} & \textbf{98.8} & \textbf{98.3} \\
\midrule
\multicolumn{11}{l}{\emph{Ours, within-subject}} \\
\quad Full sequence  & 0.262 & \textbf{0.354} & 94.9 & 98.4 & 94.5 & 92.9 & 0.647 & 0.363 & 97.2 & 97.3 \\
\quad Visual readout  & 0.260 & 0.353 & \textbf{95.3} & 98.5 & 94.7 & 93.0 & 0.642 & 0.363 & 97.2 & 97.5 \\
%\quad Visual input    & & & & & & & & & & \\
\multicolumn{11}{l}{\emph{Control, within-subject}} \\
\quad Trial permutation & 0.017 & 0.288 & 49.6 & 50.5 & 50.3 & 49.8 & 0.972 & 0.641 & 0.2 & 0.2 \\
\quad Zeroed betas  & 0.007 & 0.350 & 49.7 & 50.1 & 49.8 & 49.9 & 0.990 & 0.709 & 0.3 & - \\
\midrule
\midrule
\multicolumn{11}{l}{\emph{Ours, cross-subject}} \\
\quad Full sequence & 0.188 & 0.346 & \textbf{88.3} & \textbf{94.2} & \textbf{87.3} & \textbf{85.2} & \textbf{0.767} & \textbf{0.438}  & \textbf{69.1} & \textbf{75.5} \\
\quad Visual readout & \textbf{0.194} & \textbf{0.361} & 88.0 & 93.5 & 86.2 & 84.0 & 0.777 & 0.447 & 62.4 & 71.0 \\
\multicolumn{11}{l}{\emph{Control, cross-subject}} \\
\quad \makecell[l]{Visual-readout\\ w/o alignment} & 0.011 & 0.244 & 51.7 & 53.5 & 51.6 & 52.1 & 0.977 & 0.634 & 0.6  & 0.5 \\
\bottomrule
\end{tabular}
\end{table}

We report within-subject brain captioning and image reconstruction results in Tables~\ref{tab:cortext_caption} and~\ref{tab:image_reconstruction}, respectively (see Appendix~\ref{app:subject_results} for per-subject results and Appendix~\ref{app:image_reconstructions} for example reconstructions). Two controls remove the information decoding depends on: a \emph{trial permutation} control that breaks the correspondence between brain responses and stimuli, and a \emph{zeroed betas} control that replaces the input with zeros. Our frozen pre-trained encoders outperform both controls by a wide margin on every metric except SSIM in image reconstruction, which fails to separate the zeroed-betas control. This is consistent with \citet{mindeye2}, who note that SSIM is unreliable and can score well even for degenerate outputs. Decoding from the full parcel sequence and from the visual token subset (Section~\ref{subsec:dataset}) performs comparably on both tasks, indicating that the heads are largely insensitive to the additional, mostly non-visual parcels. Captioning is on par with the fully supervised CorText baseline; reconstruction trails MindEye2 slightly, plausibly because its head is pre-trained on seven further subjects. Together, these results show that a representation learned without access to the stimuli supports generative decoding at a level close to pipelines trained end-to-end against them.

Next, we turn to within-subject cross-trial retrieval (Table~\ref{tab:within_subject_retrieval}). Given one stimulus repetition, this evaluation tests for retrieval of another repetition of the same stimulus, measuring how well a representation preserves stimulus-related information while suppressing trial-specific variability. We first apply it to the raw betas, which serve as a control: they ablate our pre-training entirely and establish what the task yields with no learned representation at all. We then evaluate the brain embeddings learned by our objective, along with each component's contribution. The default JEPA baseline --- which draws context and target branches from the same trial and uses only the prediction loss --- falls below the raw betas. Both cross-trial pairing and the context loss are therefore necessary for stimulus-discriminative representations, with the context loss yielding the largest gain. Reliability weighting adds a small further improvement, visible mainly in mean rank, with Recall@1 essentially unchanged. Finally, restricting retrieval to the visual cortex token subset substantially improves performance over the full token sequence, raising Recall@1 from $0.39$ to $0.63$ and reducing mean rank from $31.48$ to $16.53$, indicating that the remaining parcels contribute variability that is not stimulus-locked.

\begin{table}[t]
\centering
\small
\caption{Pre-training ablation, measured by within-subject cross-trial retrieval. One repetition of a stimulus is ranked by cosine similarity against the embeddings of another repetition over the full 1000-stimulus test set, with no trained decoder; chance is mean rank $500.5$ and Recall@1 $0.001$. Results are averaged over subjects 1, 2, 5, and 7. FS reads the full sequence of 500 parcels, VS the 66-parcel visual subset selected at the encoder output. The first row uses no encoder and ranks the betas directly. Each subsequent row adds one component to the row above.}
\label{tab:within_subject_retrieval}
\begin{tabular}{lcccc}
\toprule
Method & \multicolumn{2}{c}{Mean rank$\downarrow$} & \multicolumn{2}{c}{Recall@1$\uparrow$} \\
\cmidrule(lr){2-3} \cmidrule(lr){4-5}
 & FS & VS & FS & VS \\
\midrule
Betas (no encoder) & 356.68 & 197.76 & 0.01 & 0.09 \\
\midrule
JEPA default (same-trial, $\mathcal{L}_{\text{pred}}$ only) & 392.51 & 362.69 & 0.01 & 0.01 \\
\quad + cross-trial pairing & 237.40 & 116.27 & 0.02 & 0.13 \\
\quad + context loss $\mathcal{L}_{\text{ctx}}$ & 33.28 & 18.44 & \textbf{0.39} & 0.61 \\
\quad + reliability weighting & \textbf{31.48} & \textbf{16.53} & \textbf{0.39} & \textbf{0.63} \\
\bottomrule
\end{tabular}
\end{table}

The same components that improve within-subject retrieval also facilitate cross-subject alignment (Table~\ref{tab:cross_subject_retrieval}). Here, we map an averaged response from one subject into another subject's representation space using Eq.~\ref{eq:alignment}, and retrieve the matching stimulus. We first evaluate a control operating on dimensionality-reduced betas, since fitting the unsupervised transformation in the full beta space is infeasible; it scores close to chance. Our learned representations, by contrast, support accurate cross-subject retrieval. As in the within-subject setting, cross-trial pairing and the context loss account for most of the improvement, with reliability weighting adding a small further gain. Whitening the representations after pre-training improves recall and mean rank further still, to a Recall@1 of $0.59$ at a mean rank of $7.45$. Because a separate rotation is fitted for every token, aligning the full 500-parcel sequence is considerably more expensive than aligning the 66-token visual subset, and we therefore fit it only for the complete pipeline; alignment remains possible there, reaching Recall@1 of $0.38$ at a mean rank of $17.45$. The full sequence retrieves less accurately than the visual subset by a similar margin as within subjects (Table~\ref{tab:within_subject_retrieval}), so the drop reflects the lower stimulus specificity of the mostly non-visual parcels rather than any added difficulty in aligning them. Retrieval without any learned transformation, finally, performs close to chance: independently trained subject representations are not aligned by default, yet their geometry is compatible enough that orthogonal maps recover the correspondence.

Retrieval shows that the aligned spaces correspond, but the question alignment is ultimately meant to answer is whether a decoding head can read them. We therefore transfer a head trained on one subject to another without any refinement, and evaluate it on both decoding tasks (Tables~\ref{tab:cortext_caption} and~\ref{tab:image_reconstruction}; results for all subject pairs in Appendix~\ref{app:subject_results} and example reconstructions in Appendix~\ref{app:image_reconstructions}). As a control (\emph{visual-readout w/o alignment}), we transfer the head directly between subject encoders, ablating the alignment transformation itself; performance collapses, confirming that the independently learned subject spaces are not directly interchangeable. Mapping the other subject's representations into the space on which the head was trained substantially recovers decoding performance.

Within participants, the visual cortex token subset and the full sequence perform equivalently on both tasks. Across participants, they diverge, and not in the same direction: the visual subset transfers better for captioning ($0.453$ vs.\ $0.394$ CIDEr), whereas the full sequence transfers slightly better for reconstruction on the high-level metrics ($+1.2$ CLIP, $+1.1$ Inception). Since both decoders read the same brain embeddings through the same alignment frames, the difference must lie in how they consume the token sequence. We attribute it to the pooling stage: the reconstruction head pools its input attentively and can learn which parcels to weight, whereas the captioning projector applies fixed weights to every token. Poorly aligned non-visual parcels would then be discounted in the first case and injected as noise in the second.

\begin{table}[t]
\centering
\small
\caption{Pre-training and alignment ablation, measured by cross-subject retrieval. For each pair of subjects, an orthogonal map is fitted per parcel token, from each subject's own training embeddings alone and with no shared stimuli or labels; a subject's repetition-averaged response is then mapped into the other's frame and ranked by cosine similarity against that subject's embeddings over the full 1000-stimulus test set. Chance is mean rank $500.5$ and Recall@1 $0.001$. Results are averaged over the 12 ordered pairs of subjects 1, 2, 5, and 7. The first row uses no encoder and aligns dimensionality-reduced betas directly. Each subsequent row adds one component to the row above, except the last, which leaves pre-training unchanged and whitens the embeddings after training as part of the alignment procedure. All rows use the 66-parcel visual subset selected at the encoder output.}
\label{tab:cross_subject_retrieval}
\begin{tabular}{lcc}
\toprule
Method & Mean rank$\downarrow$ & Recall@1$\uparrow$ \\
\midrule
Betas (no encoder) + PCA & 429.07 & 0.00 \\
\midrule
JEPA default (same-trial, $\mathcal{L}_{\text{pred}}$ only) & 452.40 & 0.00 \\
\quad + cross-trial pairing & 367.17 & 0.05 \\
\quad + context loss $\mathcal{L}_{\text{ctx}}$ & 15.00 & 0.45 \\
\quad + reliability weighting & 13.11 & 0.49 \\
\quad + whitening (post-training) & \textbf{7.45} & \textbf{0.59} \\
%\quad on full 500-parcel sequence & 17.91 & 0.36 \\
\bottomrule
\end{tabular}
\end{table}
\section{Conclusion}
Latent prediction offers a common representational basis for expressive within-subject decoding and for generalization across subjects. We trained one JEPA encoder per NSD participant, predicting the latent representation of masked cortical parcels from a different repetition of the same stimulus. The encoder never sees the stimulus images or any representation derived from them. Kept frozen, the encoder supports both brain captioning and image reconstruction at a level close to CorText and MindEye2, whose brain encoder is trained end-to-end on text or image targets.

Encoders trained independently for different subjects can also be aligned by orthogonal maps estimated without paired data. After alignment, a decoding head trained on one subject can be applied unchanged to another, retaining most of its within-subject quality on captioning and reconstruction alike; without alignment, decoding collapses. \citet{MarcosManchn2026PlatonicRI} showed that compact embeddings built for retrieval admit such maps. Our results extend this to representations rich enough to drive generative decoders --- which is what makes the alignment useful rather than merely measurable.

Because neither pre-training nor alignment uses the content of the stimuli, and alignment needs no stimuli shared between subjects, the approach can in principle combine data from subjects who saw different stimuli. Future work will test whether pooling data in this way improves the learned representations, whether alignment holds across datasets with different stimulus distributions, whether the objective extends to data without repeated presentations, and whether the encoders can be adapted to subjects unseen during pre-training.

% \section{Conclusion}
% We applied latent predictive training to sensory-evoked fMRI, learning a separate encoder for each subject from brain responses alone. The resulting embeddings carry the fine-grained information needed for generative neural decoding: when frozen, they support image reconstruction and brain captioning with performance comparable to pipelines trained end-to-end against the stimulus or its caption. Crucially, these embeddings can also be aligned across subjects using orthogonal transformations fitted without paired data. This alignment enables cross-subject decoding: a head trained on one subject can decode another subject's embeddings after they are mapped into the corresponding representational space, without retraining the head.

% A current limitation of our approach is its reliance on repeated stimulus presentations. Cross-trial pairing is essential for suppressing trial-specific variability and extracting stimulus-related information; without it, the learned embeddings perform worse than the raw betas in within-subject retrieval.

% Taken together, our results show that neither representation learning nor cross-subject alignment requires paired stimuli. This removes a key constraint that has limited many fMRI decoding approaches to subject-specific models and tightly matched datasets. Future work will investigate whether training on larger pools of unpaired data across subjects and datasets further improves the learned representations, and whether the encoders can adapt to subjects unseen during pre-training.

\subsection*{AI use statement}
In this work, we used generative AI tools to assist with writing. Specifically, we first drafted all sections, then used AI to refine and polish the draft. We also used AI to source relevant related work. We manually reviewed all references sourced this way. Finally, we also used AI to assist with debugging and reviewing code. We have not used generative AI tools to help develop theoretical models or conceptual frameworks, propose or refine hypotheses, design or provide feedback on research methodology or experiments, implement methods, clean and reformat datasets, support qualitative and thematic data analysis, or interpret results. The tasks of generating synthetic data sets, formulating mathematical claims, providing critical ingredients for proving mathematical claims, assisting in the writing of proofs, and assisting with translation are not applicable to this work. We take responsibility for the final content of this work, including text, claims, or artifacts produced with the aid of generative AI.

\subsection*{Ethics statement}
This work uses the Natural Scenes Dataset, a publicly released fMRI dataset collected with participant consent~\citep{allen2022massive}. We collected no new data from human participants and release no new data. Furthermore, this work builds on previous findings in brain decoding, and we do not believe any of its contributions raise specific ethical concerns.

\bibliography{iclr2027_conference}
\bibliographystyle{iclr2027_conference}

\appendix
\section{Training Parameters}\label{app:training_params}
We use the same settings for all participants across pre-training, brain captioning, image reconstruction, and alignment. During pre-training, we train on paired trials, with the context and target encoders receiving complementary single trials. During downstream training, alignment, and evaluation, we average all three repetitions per stimulus before encoding.

\begin{table}[H]
\centering
\caption{Pre-training hyperparameters for all four participants.}
\label{tab:pretrain_hparams}
\begin{tabular}{@{}p{0.42\linewidth}l@{}}
\toprule
Parameter & Value \\
\midrule
\multicolumn{2}{@{}l}{\emph{Architecture}} \\
\quad Encoder transformer depth & 2 \\
\quad Encoder attention heads & 8 \\
\quad Parcellation & Schaefer-500 \\
\quad Encoder embedding dimension & 512 \\
\quad Embedding MLP hidden dimension & 256 \\
\quad Predictor depth & 4 \\
\quad Predictor embedding dimension & 384 \\
\quad Predictor attention heads & 8 \\
\addlinespace
\multicolumn{2}{@{}l}{\emph{Masking}} \\
\quad Context blocks & 1 \\
\quad Target blocks & 4 \\
\quad Context mask scale & 0.5 -- 0.7 \\
\quad Target mask scale & 0.2 -- 0.3 \\
\addlinespace
\multicolumn{2}{@{}l}{\emph{Optimization}} \\
\quad Optimizer & AdamW \\
\quad Learning rate schedule & warmup + cosine \\
\quad Initial learning rate & $5\times10^{-5}$ \\
\quad Peak learning rate & $1\times10^{-3}$ \\
\quad Final learning rate & $1\times10^{-6}$ \\
\quad Warmup epochs & 40 \\
\quad Weight decay & 0.04 $\rightarrow$ 0.4 (cosine) \\
\quad Batch size & 512 \\
\quad Epochs & 300 \\
\quad EMA momentum & 0.996 $\rightarrow$ 1.0 \\
\bottomrule
\end{tabular}
\end{table}

\begin{table}[H]
\centering
\caption{Image reconstruction optimization parameters used within the MindEye2 pipeline.}
\label{tab:recon_hparams}
\begin{tabular}{@{}p{0.44\linewidth}l@{}}
\toprule
Parameter & Value \\
\midrule
Optimizer & AdamW \\
Learning rate schedule & cosine \\
Peak learning rate & $3\times10^{-4}$ \\
Minimum learning rate & 0 \\
Weight decay & 0.01 \\
Batch size & 64 \\
Epochs & 450 \\
\bottomrule
\end{tabular}
\end{table}

\begin{table}[t]
\centering
\caption{Brain captioning optimization parameters used within the CorText pipeline.}
\label{tab:cortext_hparams}
\begin{tabular}{@{}p{0.44\linewidth}l@{}}
\toprule
Parameter & Value \\
\midrule
Optimizer & AdamW (8-bit) \\
Learning rate schedule & cosine \\
Stage-1 learning rate & $2\times10^{-3}$ \\
Stage-2 learning rate & $2\times10^{-5}$ \\
Stage-1 weight decay & 0.05 \\
Stage-2 weight penalty & 0.001 \\
Stage-1 epochs & 20 \\
Stage-2 epochs & 5 \\
Batch size & 32 \\
LoRA rank & 16 \\
LoRA $\alpha$ & 16 \\
LoRA dropout & 0.05 \\
LoRA target modules & \texttt{q,k,v,o\_proj} \\
Quantization & 4-bit \\
\bottomrule
\end{tabular}
\end{table}

\begin{table}[t]
\centering
\caption{Cross-subject alignment hyperparameters. The map is fitted for each
ordered pair of participants from their own training embeddings alone, with no
shared stimuli, no paired samples, and no labels using the framework introduced by~\citet{MarcosManchn2026PlatonicRI}.}
\label{tab:align_hparams}
\begin{tabular}{@{}p{0.44\linewidth}l@{}}
\toprule
Parameter & Value \\
\midrule
\multicolumn{2}{@{}l}{\emph{Initialization}} \\
\quad $k$-means clusters & 40 \\
\quad Anchor restarts & 100 \\
\quad Random seeds & 5 \\
\addlinespace
\multicolumn{2}{@{}l}{\emph{Refinement}} \\
\quad ICP iterations & 100 \\
\quad ICP step size & 0.5 $\rightarrow$ 0.02 \\
\quad ICP neighbours & 20 \\
\bottomrule
\end{tabular}
\end{table}
\section{Per-Participant Results}\label{app:subject_results}
\begin{table}[H]
\centering
\small
\setlength{\tabcolsep}{2pt}
\caption{Per-subject caption generation scores from brain activity, within-subject. Higher is better for all metrics; the best result in each column and subject is shown in bold. The final block averages over subjects and corresponds to the within-subject rows of Table~\ref{tab:cortext_caption}.}
\label{tab:cortext_caption_persubj}
\begin{tabular}{llccccccccc}
\toprule
& & \multicolumn{4}{c}{BLEU} & METEOR & ROUGE-L & CIDEr & CLIP-S & RefCLIP-S \\
\cmidrule(lr){3-6}
Subject & Method & 1 & 2 & 3 & 4 & & & & & \\
\midrule
\multirow{4}{*}{1} & Full sequence & \textbf{0.630} & \textbf{0.443} & \textbf{0.310} & \textbf{0.217} & \textbf{0.201} & \textbf{0.465} & \textbf{0.689} & \textbf{0.642} & \textbf{0.711} \\
  & Visual readout & 0.622 & 0.436 & 0.301 & 0.210 & 0.199 & 0.461 & 0.681 & 0.639 & 0.709 \\
  & Trial permutation & 0.388 & 0.171 & 0.071 & 0.033 & 0.088 & 0.291 & 0.050 & 0.391 & 0.471 \\
  & Zeroed betas & 0.181 & 0.079 & 0.026 & 0.006 & 0.094 & 0.191 & 0.001 & 0.351 & 0.414 \\
\midrule
\multirow{4}{*}{2} & Full sequence & \textbf{0.629} & \textbf{0.445} & \textbf{0.311} & \textbf{0.218} & \textbf{0.205} & \textbf{0.469} & \textbf{0.683} & \textbf{0.634} & \textbf{0.704} \\
  & Visual readout & 0.614 & 0.431 & 0.294 & 0.203 & 0.197 & 0.456 & 0.647 & 0.629 & 0.699 \\
  & Trial permutation & 0.390 & 0.178 & 0.083 & 0.041 & 0.091 & 0.297 & 0.065 & 0.396 & 0.476 \\
  & Zeroed betas & 0.197 & 0.071 & 0.029 & 0.013 & 0.098 & 0.185 & 0.001 & 0.375 & 0.458 \\
\midrule
\multirow{4}{*}{5} & Full sequence & 0.631 & 0.439 & 0.297 & 0.204 & 0.205 & 0.462 & 0.702 & 0.652 & 0.720 \\
  & Visual readout & \textbf{0.653} & \textbf{0.472} & \textbf{0.332} & \textbf{0.237} & \textbf{0.215} & \textbf{0.487} & \textbf{0.771} & \textbf{0.656} & \textbf{0.725} \\
  & Trial permutation & 0.392 & 0.178 & 0.076 & 0.035 & 0.091 & 0.296 & 0.057 & 0.395 & 0.475 \\
  & Zeroed betas & 0.185 & 0.076 & 0.025 & 0.010 & 0.090 & 0.192 & 0.002 & 0.363 & 0.432 \\
\midrule
\multirow{4}{*}{7} & Full sequence & \textbf{0.612} & \textbf{0.426} & \textbf{0.295} & \textbf{0.210} & \textbf{0.195} & \textbf{0.459} & \textbf{0.617} & \textbf{0.611} & \textbf{0.683} \\
  & Visual readout & 0.604 & 0.418 & 0.287 & 0.200 & 0.190 & 0.450 & 0.598 & 0.608 & 0.680 \\
  & Trial permutation & 0.394 & 0.179 & 0.084 & 0.045 & 0.091 & 0.298 & 0.070 & 0.393 & 0.473 \\
  & Zeroed betas & 0.196 & 0.084 & 0.032 & 0.011 & 0.096 & 0.204 & 0.001 & 0.367 & 0.407 \\
\midrule
\multirow{4}{*}{\emph{Mean}} & Full sequence & \textbf{0.625} & 0.438 & 0.303 & 0.212 & \textbf{0.201} & \textbf{0.464} & 0.673 & \textbf{0.635} & \textbf{0.705} \\
            & Visual readout & 0.623 & \textbf{0.439} & \textbf{0.304} & \textbf{0.212} & 0.200 & 0.463 & \textbf{0.674} & 0.633 & 0.703 \\
            & Trial permutation & 0.391 & 0.176 & 0.078 & 0.039 & 0.090 & 0.296 & 0.061 & 0.394 & 0.474 \\
            & Zeroed betas & 0.190 & 0.078 & 0.028 & 0.010 & 0.094 & 0.193 & 0.001 & 0.364 & 0.428 \\
\bottomrule
\end{tabular}
\end{table}

\begin{table}[t]
\centering
\small
\setlength{\tabcolsep}{3pt}
\caption{Per-pair caption generation from brain activity, cross-subject. A head trained on one subject is applied to the aligned representations of another. Higher is better for all metrics; the best result in each column and pair is shown in bold. The final block averages over the 12 ordered pairs and corresponds to the cross-subject rows of Table~\ref{tab:cortext_caption}.}
\label{tab:cortext_caption_perpair}
\begin{tabular}{llccccccccc}
\toprule
& & \multicolumn{4}{c}{BLEU} & METEOR & ROUGE-L & CIDEr & CLIP-S & RefCLIP-S \\
\cmidrule(lr){3-6}
Pair & Method & 1 & 2 & 3 & 4 & & & & & \\
\midrule
\multirow{3}{*}{$2 \rightarrow 1$} & Full sequence & 0.526 & 0.337 & 0.208 & 0.133 & 0.160 & 0.403 & 0.392 & 0.573 & 0.644 \\
                  & Visual readout & \textbf{0.557} & \textbf{0.358} & \textbf{0.228} & \textbf{0.153} & \textbf{0.169} & \textbf{0.421} & \textbf{0.468} & \textbf{0.583} & \textbf{0.656} \\
                  & w/o alignment & 0.002 & 0.000 & 0.000 & 0.000 & 0.001 & 0.002 & 0.000 & 0.532 & 0.591 \\
\midrule
\multirow{3}{*}{$5 \rightarrow 1$} & Full sequence & 0.574 & 0.383 & \textbf{0.252} & \textbf{0.172} & 0.180 & 0.432 & \textbf{0.543} & 0.605 & 0.675 \\
                  & Visual readout & \textbf{0.579} & \textbf{0.383} & 0.252 & 0.169 & \textbf{0.180} & \textbf{0.434} & 0.542 & \textbf{0.608} & \textbf{0.679} \\
                  & w/o alignment & 0.001 & 0.000 & 0.000 & 0.000 & 0.000 & 0.001 & 0.000 & 0.534 & 0.593 \\
\midrule
\multirow{3}{*}{$7 \rightarrow 1$} & Full sequence & 0.449 & 0.268 & 0.155 & 0.095 & 0.145 & 0.372 & 0.301 & 0.549 & 0.622 \\
                  & Visual readout & \textbf{0.518} & \textbf{0.330} & \textbf{0.210} & \textbf{0.139} & \textbf{0.158} & \textbf{0.399} & \textbf{0.380} & \textbf{0.552} & \textbf{0.626} \\
                  & w/o alignment & 0.013 & 0.000 & 0.000 & 0.000 & 0.007 & 0.017 & 0.000 & 0.509 & 0.569 \\
\midrule
\multirow{3}{*}{$1 \rightarrow 2$} & Full sequence & 0.471 & 0.302 & 0.189 & 0.122 & 0.169 & 0.401 & 0.429 & 0.583 & 0.656 \\
                  & Visual readout & \textbf{0.582} & \textbf{0.397} & \textbf{0.268} & \textbf{0.183} & \textbf{0.181} & \textbf{0.438} & \textbf{0.565} & \textbf{0.606} & \textbf{0.678} \\
                  & w/o alignment & 0.049 & 0.009 & 0.000 & 0.000 & 0.027 & 0.062 & 0.000 & 0.473 & 0.542 \\
\midrule
\multirow{3}{*}{$5 \rightarrow 2$} & Full sequence & 0.569 & 0.380 & 0.250 & 0.172 & 0.181 & 0.433 & 0.548 & \textbf{0.607} & \textbf{0.679} \\
                  & Visual readout & \textbf{0.591} & \textbf{0.408} & \textbf{0.280} & \textbf{0.195} & \textbf{0.186} & \textbf{0.449} & \textbf{0.570} & 0.604 & 0.675 \\
                  & w/o alignment & 0.059 & 0.014 & 0.000 & 0.000 & 0.030 & 0.078 & 0.000 & 0.472 & 0.539 \\
\midrule
\multirow{3}{*}{$7 \rightarrow 2$} & Full sequence & 0.485 & 0.300 & 0.184 & 0.119 & 0.149 & 0.387 & 0.345 & 0.542 & 0.616 \\
                  & Visual readout & \textbf{0.522} & \textbf{0.334} & \textbf{0.210} & \textbf{0.136} & \textbf{0.158} & \textbf{0.401} & \textbf{0.396} & \textbf{0.564} & \textbf{0.637} \\
                  & w/o alignment & 0.058 & 0.012 & 0.000 & 0.000 & 0.031 & 0.075 & 0.000 & 0.487 & 0.551 \\
\midrule
\multirow{3}{*}{$1 \rightarrow 5$} & Full sequence & 0.473 & 0.298 & 0.181 & 0.112 & 0.157 & 0.387 & 0.373 & 0.566 & 0.637 \\
                  & Visual readout & \textbf{0.544} & \textbf{0.356} & \textbf{0.229} & \textbf{0.150} & \textbf{0.167} & \textbf{0.412} & \textbf{0.438} & \textbf{0.583} & \textbf{0.656} \\
                  & w/o alignment & 0.052 & 0.011 & 0.002 & 0.000 & 0.030 & 0.070 & 0.000 & 0.482 & 0.550 \\
\midrule
\multirow{3}{*}{$2 \rightarrow 5$} & Full sequence & 0.417 & 0.255 & 0.147 & 0.088 & 0.145 & 0.367 & 0.295 & \textbf{0.555} & 0.625 \\
                  & Visual readout & \textbf{0.477} & \textbf{0.290} & \textbf{0.171} & \textbf{0.103} & \textbf{0.149} & \textbf{0.378} & \textbf{0.312} & 0.554 & \textbf{0.626} \\
                  & w/o alignment & 0.049 & 0.009 & 0.000 & 0.000 & 0.026 & 0.065 & 0.000 & 0.493 & 0.556 \\
\midrule
\multirow{3}{*}{$7 \rightarrow 5$} & Full sequence & 0.405 & 0.237 & 0.138 & 0.085 & 0.142 & 0.357 & 0.281 & 0.530 & 0.599 \\
                  & Visual readout & \textbf{0.507} & \textbf{0.317} & \textbf{0.195} & \textbf{0.125} & \textbf{0.153} & \textbf{0.394} & \textbf{0.317} & \textbf{0.543} & \textbf{0.615} \\
                  & w/o alignment & 0.052 & 0.009 & 0.000 & 0.000 & 0.026 & 0.071 & 0.000 & 0.474 & 0.542 \\
\midrule
\multirow{3}{*}{$1 \rightarrow 7$} & Full sequence & 0.535 & 0.332 & 0.203 & 0.126 & 0.155 & 0.396 & 0.346 & 0.555 & 0.628 \\
                  & Visual readout & \textbf{0.566} & \textbf{0.378} & \textbf{0.250} & \textbf{0.168} & \textbf{0.174} & \textbf{0.423} & \textbf{0.475} & \textbf{0.583} & \textbf{0.656} \\
                  & w/o alignment & 0.048 & 0.000 & 0.000 & 0.000 & 0.010 & 0.045 & 0.002 & 0.528 & 0.597 \\
\midrule
\multirow{3}{*}{$2 \rightarrow 7$} & Full sequence & 0.558 & 0.364 & 0.236 & 0.154 & 0.167 & 0.412 & 0.456 & 0.579 & 0.651 \\
                  & Visual readout & \textbf{0.571} & \textbf{0.383} & \textbf{0.254} & \textbf{0.174} & \textbf{0.172} & \textbf{0.422} & \textbf{0.485} & \textbf{0.581} & \textbf{0.653} \\
                  & w/o alignment & 0.038 & 0.005 & 0.000 & 0.000 & 0.008 & 0.032 & 0.001 & 0.528 & 0.594 \\
\midrule
\multirow{3}{*}{$5 \rightarrow 7$} & Full sequence & 0.536 & 0.347 & 0.221 & 0.146 & 0.163 & 0.408 & 0.412 & 0.571 & 0.643 \\
                  & Visual readout & \textbf{0.572} & \textbf{0.376} & \textbf{0.247} & \textbf{0.165} & \textbf{0.175} & \textbf{0.426} & \textbf{0.483} & \textbf{0.584} & \textbf{0.657} \\
                  & w/o alignment & 0.029 & 0.004 & 0.000 & 0.000 & 0.007 & 0.027 & 0.001 & 0.529 & 0.599 \\
\midrule
\multirow{3}{*}{\emph{Mean}} & Full sequence & 0.500 & 0.317 & 0.197 & 0.127 & 0.159 & 0.396 & 0.393 & 0.568 & 0.639 \\
            & Visual readout & \textbf{0.549} & \textbf{0.359} & \textbf{0.233} & \textbf{0.155} & \textbf{0.168} & \textbf{0.416} & \textbf{0.453} & \textbf{0.579} & \textbf{0.651} \\
            & w/o alignment & 0.038 & 0.006 & 0.000 & 0.000 & 0.017 & 0.046 & 0.000 & 0.503 & 0.569 \\
\bottomrule
\end{tabular}
\end{table}

\begin{table}[t]
\centering
\small
\setlength{\tabcolsep}{1pt}
\caption{Per-subject image reconstruction and retrieval, within-subject. Arrows indicate the direction of improvement; the best result in each column and subject is shown in bold. The final block averages over subjects and corresponds to the within-subject rows of Table~\ref{tab:image_reconstruction}. Brain retrieval is undefined for the zeroed-betas control and is omitted.}
\label{tab:image_reconstruction_persubj}
\begin{tabular}{llcccccccccc}
\toprule
& & \multicolumn{4}{c}{Low-Level} & \multicolumn{4}{c}{High-Level} & \multicolumn{2}{c}{Retrieval} \\
\cmidrule(lr){3-6} \cmidrule(lr){7-10} \cmidrule(lr){11-12}
Subject & Method & PixCorr$\uparrow$ & SSIM$\uparrow$ & Alex(2)$\uparrow$ & Alex(5)$\uparrow$
 & Incep$\uparrow$ & CLIP$\uparrow$ & Eff$\downarrow$ & SwAV$\downarrow$ & Image$\uparrow$ & Brain$\uparrow$ \\
\midrule
\multirow{4}{*}{1} & Full sequence & 0.309 & \textbf{0.369} & \textbf{97.3} & 99.2 & 95.3 & 93.7 & 0.631 & 0.352 & 99.9 & \textbf{100.0} \\
 & Visual readout & \textbf{0.313} & 0.364 & 97.2 & \textbf{99.2} & \textbf{95.7} & \textbf{93.8} & \textbf{0.625} & \textbf{0.349} & \textbf{99.9} & 99.9 \\
 & Trial permutation & 0.016 & 0.286 & 48.1 & 49.3 & 49.8 & 49.6 & 0.973 & 0.643 & 0.2 & 0.2 \\
 & Zeroed betas & 0.018 & 0.255 & 49.8 & 50.9 & 49.7 & 49.7 & 0.977 & 0.700 & 0.4 & -- \\
\midrule
\multirow{4}{*}{2} & Full sequence & 0.276 & 0.357 & 96.8 & 99.2 & 95.4 & 93.1 & 0.632 & \textbf{0.351} & \textbf{99.9} & \textbf{99.9} \\
 & Visual readout & \textbf{0.277} & 0.361 & \textbf{97.1} & \textbf{99.3} & \textbf{95.5} & \textbf{94.0} & \textbf{0.628} & 0.352 & 99.9 & 99.9 \\
 & Trial permutation & 0.013 & 0.291 & 49.4 & 50.3 & 49.8 & 49.8 & 0.970 & 0.641 & 0.1 & 0.2 \\
 & Zeroed betas & 0.067 & \textbf{0.375} & 49.7 & 49.9 & 50.3 & 49.9 & 0.992 & 0.742 & 0.3 & -- \\
\midrule
\multirow{4}{*}{5} & Full sequence & 0.233 & 0.345 & 93.6 & 98.3 & 95.5 & 94.0 & 0.626 & 0.357 & \textbf{96.7} & 97.0 \\
 & Visual readout & \textbf{0.237} & 0.348 & \textbf{94.9} & \textbf{98.6} & \textbf{95.7} & \textbf{94.2} & \textbf{0.625} & \textbf{0.356} & 96.6 & \textbf{97.2} \\
 & Trial permutation & 0.021 & 0.291 & 50.9 & 51.5 & 50.0 & 50.3 & 0.971 & 0.640 & 0.2 & 0.3 \\
 & Zeroed betas & -0.060 & \textbf{0.428} & 49.7 & 50.1 & 49.0 & 49.8 & 0.998 & 0.708 & 0.3 & -- \\
\midrule
\multirow{4}{*}{7} & Full sequence & \textbf{0.230} & \textbf{0.345} & 91.9 & \textbf{96.8} & 91.8 & \textbf{90.6} & 0.700 & 0.394 & 92.1 & 92.3 \\
 & Visual readout & 0.214 & 0.340 & \textbf{91.9} & 96.7 & \textbf{92.1} & 90.2 & \textbf{0.692} & \textbf{0.393} & \textbf{92.6} & \textbf{93.0} \\
 & Trial permutation & 0.018 & 0.283 & 49.8 & 50.9 & 51.6 & 49.6 & 0.972 & 0.640 & 0.1 & 0.2 \\
 & Zeroed betas & 0.002 & 0.342 & 49.8 & 49.6 & 50.0 & 50.1 & 0.992 & 0.685 & 0.3 & -- \\
\midrule
\multirow{4}{*}{\emph{Mean}} & Full sequence & \textbf{0.262} & \textbf{0.354} & 94.9 & 98.4 & 94.5 & 92.9 & 0.647 & 0.363 & 97.2 & 97.3 \\
 & Visual readout & 0.260 & 0.353 & \textbf{95.3} & \textbf{98.5} & \textbf{94.7} & \textbf{93.0} & \textbf{0.642} & \textbf{0.363} & \textbf{97.2} & \textbf{97.5} \\
 & Trial permutation & 0.017 & 0.288 & 49.6 & 50.5 & 50.3 & 49.8 & 0.972 & 0.641 & 0.2 & 0.2 \\
 & Zeroed betas & 0.007 & 0.350 & 49.7 & 50.1 & 49.8 & 49.9 & 0.990 & 0.709 & 0.3 & -- \\
\bottomrule
\end{tabular}
\end{table}

\begin{table}[t]
\centering
\footnotesize
\setlength{\tabcolsep}{1.5pt}
\caption{Per-pair image reconstruction and retrieval, cross-subject. A head trained on one subject is applied to the aligned representations of another. Arrows indicate the direction of improvement; the best result in each column and pair is shown in bold. The final block averages over the 12 ordered pairs and corresponds to the cross-subject rows
of Table~\ref{tab:image_reconstruction}.}
\label{tab:image_reconstruction_perpair}
\begin{tabular}{llcccccccccc}
\toprule
& & \multicolumn{4}{c}{Low-Level} & \multicolumn{4}{c}{High-Level} & \multicolumn{2}{c}{Retrieval} \\
\cmidrule(lr){3-6} \cmidrule(lr){7-10} \cmidrule(lr){11-12}
Pair & Method & PixCorr$\uparrow$ & SSIM$\uparrow$ & Alex(2)$\uparrow$ & Alex(5)$\uparrow$
 & Incep$\uparrow$ & CLIP$\uparrow$ & Eff$\downarrow$ & SwAV$\downarrow$ & Image$\uparrow$ & Brain$\uparrow$ \\
\midrule
\multirow{3}{*}{$2 \rightarrow 1$} & Full sequence & 0.235 & 0.355 & \textbf{93.8} & \textbf{97.4} & \textbf{90.0} & \textbf{87.5} & \textbf{0.731} & \textbf{0.405} & \textbf{94.0} & \textbf{97.0} \\
 & Visual readout & \textbf{0.242} & \textbf{0.382} & 91.8 & 96.0 & 87.6 & 85.2 & 0.759 & 0.435 & 79.4 & 90.0 \\
 & w/o alignment & -0.008 & 0.213 & 55.4 & 59.4 & 54.1 & 54.5 & 0.972 & 0.627 & 0.8 & 0.8 \\
\midrule
\multirow{3}{*}{$5 \rightarrow 1$} & Full sequence & 0.172 & 0.339 & \textbf{87.7} & \textbf{94.7} & \textbf{90.4} & \textbf{88.0} & \textbf{0.739} & \textbf{0.419} & \textbf{57.1} & \textbf{62.4} \\
 & Visual readout & \textbf{0.205} & \textbf{0.374} & 87.1 & 93.2 & 88.8 & 86.6 & 0.753 & 0.436 & 35.0 & 49.4 \\
 & w/o alignment & -0.004 & 0.218 & 49.1 & 49.6 & 49.0 & 51.4 & 0.983 & 0.642 & 0.2 & 0.2 \\
\midrule
\multirow{3}{*}{$7 \rightarrow 1$} & Full sequence & 0.165 & 0.319 & \textbf{84.9} & \textbf{90.4} & \textbf{82.9} & \textbf{80.5} & \textbf{0.815} & \textbf{0.470} & \textbf{53.1} & \textbf{63.2} \\
 & Visual readout & \textbf{0.188} & \textbf{0.363} & 83.4 & 88.3 & 81.4 & 79.5 & 0.827 & 0.485 & 32.9 & 47.6 \\
 & w/o alignment & -0.005 & 0.182 & 48.8 & 48.4 & 49.9 & 50.9 & 0.982 & 0.643 & 0.4 & 0.5 \\
\midrule
\multirow{3}{*}{$1 \rightarrow 2$} & Full sequence & \textbf{0.240} & \textbf{0.370} & 93.5 & \textbf{97.5} & \textbf{90.5} & \textbf{89.3} & \textbf{0.726} & \textbf{0.410} & \textbf{95.5} & \textbf{98.0} \\
 & Visual readout & 0.232 & 0.357 & \textbf{93.6} & 96.8 & 89.6 & 86.9 & 0.739 & 0.425 & 94.9 & 97.0 \\
 & w/o alignment & -0.000 & 0.234 & 53.9 & 53.8 & 50.0 & 50.8 & 0.986 & 0.647 & 0.9 & 0.7 \\
\midrule
\multirow{3}{*}{$5 \rightarrow 2$} & Full sequence & \textbf{0.179} & \textbf{0.348} & \textbf{87.8} & \textbf{95.0} & \textbf{91.0} & \textbf{88.1} & \textbf{0.726} & \textbf{0.415} & \textbf{57.4} & \textbf{61.7} \\
 & Visual readout & 0.174 & 0.345 & 87.3 & 94.3 & 89.2 & 86.4 & 0.752 & 0.431 & 57.1 & 58.8 \\
 & w/o alignment & 0.003 & 0.207 & 51.3 & 57.7 & 56.6 & 56.5 & 0.962 & 0.616 & 0.6 & 0.6 \\
\midrule
\multirow{3}{*}{$7 \rightarrow 2$} & Full sequence & \textbf{0.167} & 0.335 & 84.8 & \textbf{91.8} & \textbf{83.8} & \textbf{82.2} & \textbf{0.803} & \textbf{0.461} & \textbf{54.2} & \textbf{62.2} \\
 & Visual readout & 0.164 & \textbf{0.338} & \textbf{85.1} & 91.2 & 82.8 & 80.3 & 0.820 & 0.474 & 52.0 & 58.3 \\
 & w/o alignment & 0.014 & 0.254 & 51.2 & 51.0 & 51.4 & 51.1 & 0.976 & 0.617 & 0.8 & 0.6 \\
\midrule
\multirow{3}{*}{$1 \rightarrow 5$} & Full sequence & 0.191 & 0.354 & \textbf{89.3} & 94.6 & 86.5 & 84.4 & 0.770 & 0.436 & \textbf{74.0} & \textbf{81.3} \\
 & Visual readout & \textbf{0.212} & \textbf{0.370} & 88.7 & \textbf{94.7} & \textbf{87.3} & \textbf{85.0} & \textbf{0.765} & \textbf{0.435} & 68.8 & 78.3 \\
 & w/o alignment & 0.041 & 0.270 & 56.5 & 61.2 & 54.2 & 55.5 & 0.961 & 0.606 & 0.3 & 0.4 \\
\midrule
\multirow{3}{*}{$2 \rightarrow 5$} & Full sequence & 0.170 & 0.352 & \textbf{88.3} & \textbf{94.4} & \textbf{87.7} & \textbf{84.4} & \textbf{0.761} & \textbf{0.437} & \textbf{74.9} & \textbf{82.3} \\
 & Visual readout & \textbf{0.189} & \textbf{0.371} & 87.5 & 94.0 & 87.0 & 82.8 & 0.776 & 0.443 & 68.8 & 77.8 \\
 & w/o alignment & 0.053 & 0.280 & 57.2 & 63.0 & 56.7 & 56.9 & 0.962 & 0.612 & 0.7 & 0.5 \\
\midrule
\multirow{3}{*}{$7 \rightarrow 5$} & Full sequence & 0.165 & 0.336 & \textbf{84.6} & \textbf{90.7} & \textbf{83.8} & \textbf{81.3} & \textbf{0.819} & \textbf{0.471} & \textbf{58.3} & \textbf{67.6} \\
 & Visual readout & \textbf{0.165} & \textbf{0.363} & 83.6 & 90.5 & 82.2 & 79.3 & 0.822 & 0.474 & 52.1 & 63.6 \\
 & w/o alignment & -0.008 & 0.275 & 47.9 & 47.0 & 50.6 & 50.0 & 0.982 & 0.650 & 0.3 & 0.4 \\
\midrule
\multirow{3}{*}{$1 \rightarrow 7$} & Full sequence & \textbf{0.196} & 0.349 & 88.4 & \textbf{95.0} & 85.9 & 84.7 & 0.778 & 0.447 & 75.6 & 80.5 \\
 & Visual readout & 0.189 & \textbf{0.357} & \textbf{89.7} & 94.6 & \textbf{86.6} & \textbf{84.9} & \textbf{0.771} & \textbf{0.440} & \textbf{76.3} & \textbf{83.0} \\
 & w/o alignment & 0.045 & 0.240 & 48.7 & 49.5 & 48.1 & 48.9 & 0.992 & 0.672 & 1.0 & 0.9 \\
\midrule
\multirow{3}{*}{$2 \rightarrow 7$} & Full sequence & 0.182 & 0.351 & 88.3 & \textbf{94.6} & \textbf{86.7} & \textbf{85.0} & \textbf{0.774} & \textbf{0.442} & \textbf{73.3} & 79.4 \\
 & Visual readout & \textbf{0.188} & \textbf{0.359} & \textbf{90.2} & 94.3 & 85.2 & 84.3 & 0.781 & 0.446 & 71.9 & \textbf{80.5} \\
 & w/o alignment & 0.019 & 0.254 & 49.0 & 51.9 & 49.3 & 49.2 & 0.983 & 0.644 & 0.5 & 0.4 \\
\midrule
\multirow{3}{*}{$5 \rightarrow 7$} & Full sequence & \textbf{0.190} & 0.340 & 87.6 & \textbf{94.7} & \textbf{88.0} & \textbf{87.0} & \textbf{0.761} & 0.438 & \textbf{62.1} & \textbf{70.0} \\
 & Visual readout & 0.178 & \textbf{0.353} & \textbf{87.7} & 93.8 & 87.0 & 86.2 & 0.765 & \textbf{0.435} & 59.4 & 67.4 \\
 & w/o alignment & -0.018 & 0.301 & 51.2 & 49.3 & 49.1 & 50.0 & 0.982 & 0.636 & 0.4 & 0.4 \\
\midrule
\multirow{3}{*}{\emph{Mean}} & Full sequence & 0.188 & 0.346 & \textbf{88.3} & \textbf{94.2} & \textbf{87.3} & \textbf{85.2} & \textbf{0.767} & \textbf{0.438} & \textbf{69.1} & \textbf{75.5} \\
 & Visual readout & \textbf{0.194} & \textbf{0.361} & 88.0 & 93.5 & 86.2 & 84.0 & 0.777 & 0.447 & 62.4 & 71.0 \\
 & w/o alignment & 0.011 & 0.244 & 51.7 & 53.5 & 51.6 & 52.1 & 0.977 & 0.634 & 0.6 & 0.5 \\
\bottomrule
\end{tabular}
\end{table}
\section{Image Reconstructions}\label{app:image_reconstructions}
\begin{figure*}[ht]
\centering
\includegraphics[width=\textwidth]{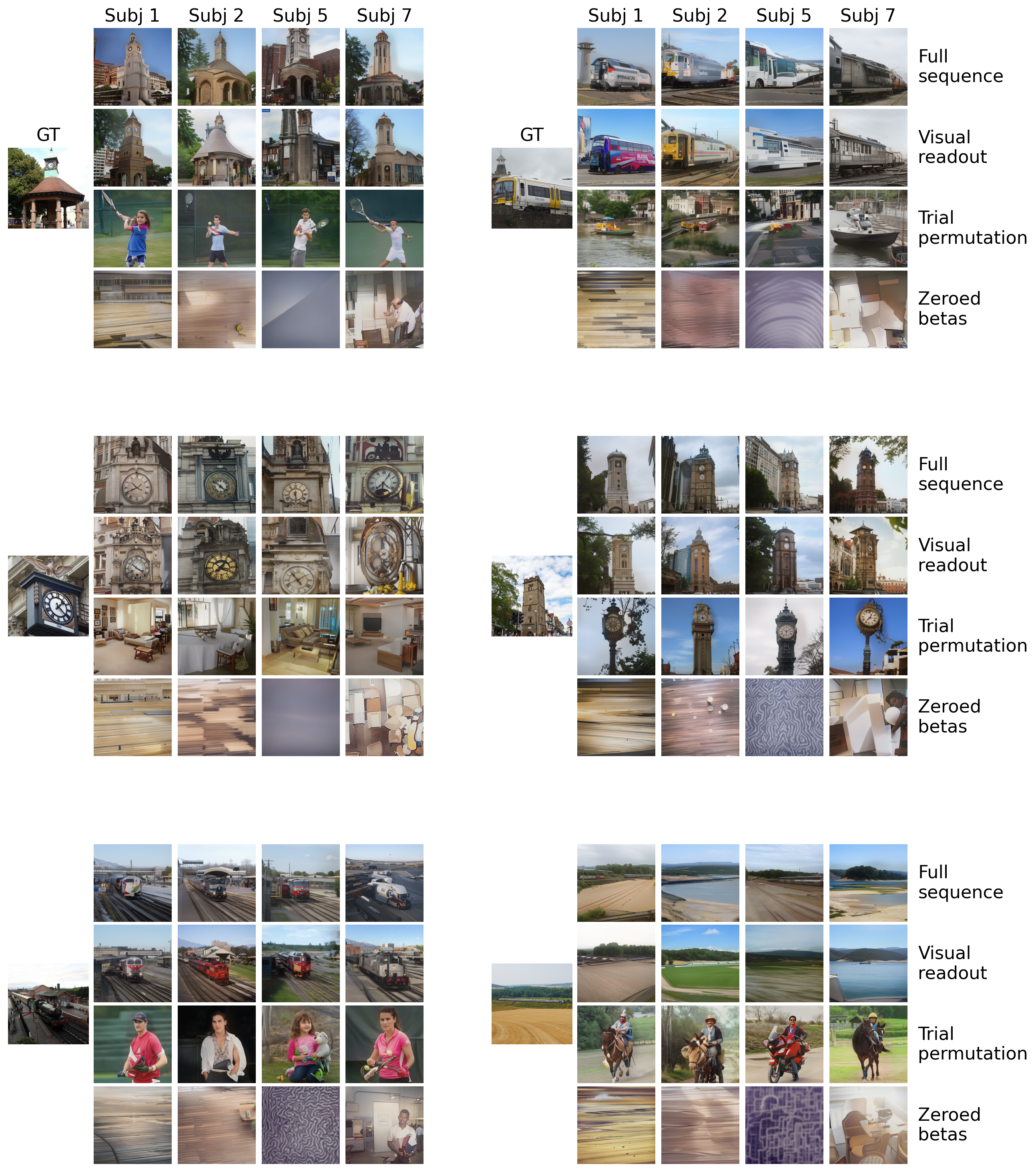}
\caption{Image reconstruction from brain activity, within-subject. Reconstructions use the refined MindEye2 pipeline, whereas Table~\ref{tab:image_reconstruction} reports metrics on the unrefined outputs. \emph{Full sequence}, \emph{Visual readout} and the control conditions are defined as in Table~\ref{tab:cortext_caption}. Stimuli were drawn at random from the 1000-image test set with a fixed seed.}
\label{fig:recon_within}
\end{figure*}

\begin{figure*}[ht]
\centering
\includegraphics[width=\textwidth]{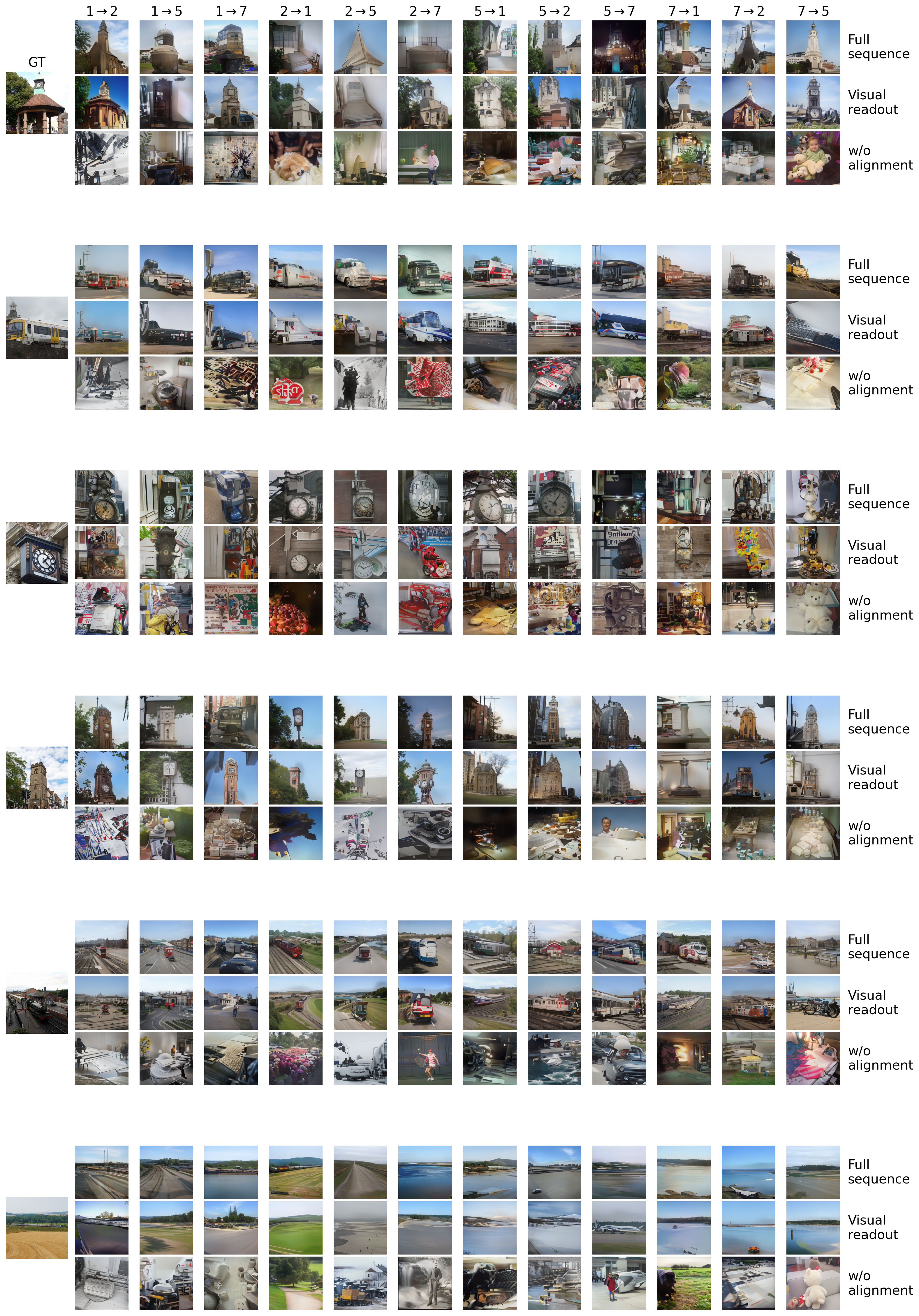}
\caption{Image reconstruction from brain activity, cross-subject. Reconstructions use the refined MindEye2 pipeline, whereas Table~\ref{tab:image_reconstruction} reports metrics on the unrefined outputs. The \emph{w/o alignment} control transfers the head between encoders without the alignment transformation. The same stimuli are shown as in Figure~\ref{fig:recon_within}.}
\label{fig:recon_cross}
\end{figure*}

\end{document}